\documentclass[10pt,twocolumn,letterpaper]{article}

\usepackage{iccv}              % To produce the CAMERA-READY version
\PassOptionsToPackage{table,dvipsnames}{xcolor}
\usepackage{xcolor}
\usepackage{colortbl}
\definecolor{iccvblue}{rgb}{0.21,0.49,0.74}
\usepackage[colorlinks=true, citecolor=iccvblue, linkcolor=red, urlcolor=red]{hyperref}
\usepackage{makecell}
\usepackage{multirow}
\usepackage{booktabs}
\usepackage{fontawesome5}
\usepackage{pifont}
\usepackage{stix}

\definecolor{lit-org}{rgb}{0.95, 0.72, 0.35}
\definecolor{lit-red}{rgb}{1, 0.449, 0.449}
\definecolor{lit-blue}{rgb}{0.449, 0.727, 0.88}

\newcommand{\J}{$\mathcal{J}$\xspace}
\newcommand{\F}{$\mathcal{F}$\xspace}
\newcommand{\JF}{$\mathcal{J}\&\mathcal{F}$\xspace}

\newcommand{\Fo}{$\dot{\mathcal{F}}$\xspace}
\newcommand{\JFo}{$\mathcal{J}\&\dot{\mathcal{F}}$\xspace}
\newcommand{\JFnd}{$\mathcal{J}\&\dot{\mathcal{F}}_d$\xspace}
\newcommand{\JFnr}{$\mathcal{J}\&\dot{\mathcal{F}}_r$\xspace}

\def\paperID{*****} % *** Enter the Paper ID here
\def\confName{ICCV}
\def\confYear{2025}

\title{Re-Prompting SAM 3 via Object Retrieval: 3rd of the 5th PVUW MOSE Track}

\author{
    Mingqi Gao\textsuperscript{1}\quad
    Sijie Li\textsuperscript{1}\quad
    Jungong Han\textsuperscript{2}\footnotemark[1] \\
 \textsuperscript{1} \normalsize{School of Computer Science, University of Sheffield} \;
 \textsuperscript{2} \normalsize{Department of Automation, Tsinghua University} \;\\
  {\tt\normalsize $\{$m.gao,sli256$\}$@sheffield.ac.uk, jghan@tsinghua.edu.cn}
}

\begin{document}
\maketitle
\renewcommand{\thefootnote}{\fnsymbol{footnote}}
\footnotetext[1]{Corresponding author.}
\begin{abstract}
This technical report explores the MOSEv2 track of the PVUW 2026 Challenge, which targets complex semi-supervised video object segmentation. Built on SAM~3, we develop an automatic re-prompting framework to improve robustness under target disappearance and reappearance, severe transformation, and strong same-category distractors. Our method first applies the SAM~3 detector to later frames to identify same-category object candidates, and then performs DINOv3-based object-level matching with a transformation-aware target feature pool to retrieve reliable target anchors. These anchors are injected back into the SAM~3 tracker together with the first-frame mask, enabling multi-anchor propagation rather than relying solely on the initial prompt. This simple directly benefits several core challenges of MOSEv2. Our solution achieves a \JFo of 51.17\% on the test set, ranking 3rd in the MOSEv2 track.
\end{abstract}    
\section{Introduction}
\label{sec:intro}

The Pixel-level Video Understanding in the Wild (PVUW) workshop has become an important platform for advancing pixel-level video understanding toward realistic scenarios. By providing large-scale benchmarks and evaluation protocols that are closer to real applications, including MOSEv2~\cite{ding2025mosev2} and MeViSv2~\cite{ding2025mevis}, PVUW promotes the study of video understanding under diverse inputs and challenging real-world conditions. With the newly introduced Audio-based Referring Video Object Segmentation task this year, PVUW now covers three highly challenging settings, namely RGB-only understanding with MOSEv2, RGB+Text understanding with MeViSv2-Text, and RGB+Audio understanding with MeViSv2-Audio. This benchmark suite further broadens the scope of pixel-level video understanding in practical scenarios and encourages the community to move toward solving more general and more complex problems.

Our submission focuses on MOSEv2~\cite{ding2025mosev2}, which belongs to the semi-supervised video object segmentation (VOS) setting. In this task, the segmentation mask of the target object is provided in the first frame, and the goal is to segment the same object throughout the remaining video frames~\cite{gao2023deep}. Semi-supervised VOS is a fundamental problem in video understanding and has broad value for real applications. Accurate and temporally consistent object segmentation can benefit emerging applications such as controllable video editing~\cite{wang2025videodirector} and embodied AI systems that require robust object-centric perception in dynamic environments~\cite{chen2025m}.

To promote methods that better support real applications, many benchmarks have been introduced over the past years. Compared with earlier datasets such as DAVIS~\cite{perazzi2016benchmark} and the YouTube-VOS series~\cite{xu2018youtube}, the MOSE (coMplex video Object SEgmentation) benchmarks~\cite{ding2023mose,ding2025mosev2} introduce challenges that are closer to realistic video understanding scenarios, including frequent occlusions, severe interference from many same-category objects, and longer temporal contexts. Compared with MOSEv1~\cite{ding2023mose}, MOSEv2~\cite{ding2025mosev2}, which is used in this challenge, further increases both the severity and frequency of occlusions and expands the diversity of target variations, including drastic scale changes, lightening transitions, and large changes in object shape, pose, and orientation. It also includes more cases where successful tracking requires open knowledge and higher-level video understanding to distinguish the target from semantically similar distractors, making the benchmark closer to real-world video perception. In addition, MOSEv2 introduces updated evaluation metrics that provide more accurate assessment of fine-grained boundaries and small-object segmentation quality.

Most existing VOS approaches rely on memory-based designs~\cite{oh2019video}. The core idea is to store segmentation results from past frames as memory, and then segment the current frame by attending to the stored memory together with the current-frame features. Compared with using only the first frame as reference, maintaining multiple past frames allows the model to preserve more target states over time, thus improving adaptation to temporal appearance changes. Following the remarkable success of SAM in image segmentation~\cite{kirillov2023segment}, its video extension SAM~2~\cite{ravi2024sam2} brought this strong segmentation capability into the video domain through an explicit memory mechanism. Due to its concise architecture and strong performance, many subsequent methods have focused on improving memory updates. For example, SAM2Long~\cite{ding2024sam2long} introduces a tree-based memory update strategy, SAMURAI~\cite{yang2024samurai} incorporates motion-aware memory updates, DAM4SAM~\cite{videnovic2025distractor} proposes distractor-aware memory design, and SeC~\cite{zhang2025sec} further augments fine-grained memory with concept-aware cues from LVLMs. More recently, SAM~3~\cite{sam3} extends the SAM~2 tracker with a parallel detector branch, enabling category-level detection and segmentation based on either text prompts or visual prompts such as points and boxes. By combining the detector outputs with tracker propagation, SAM~3 substantially alleviates the gradual degradation of segmentation quality over time and further expands the framework toward language-guided video understanding.

Despite these advances, existing methods still largely follow the conventional VOS inference setting, where only the first frame is treated as the visual anchor, and the remaining frames are segmented mainly by measuring similarity to memory frames. Under the frequent disappearance and reappearance patterns in MOSEv2, this design remains vulnerable to severe interference from similar-category objects, and the predicted mask can gradually drift to distractors as the video proceeds. To address this issue, we propose to move beyond single-frame anchoring and introduce additional anchors from later frames to improve robustness against large appearance changes and long-term ambiguity. Specifically, before the main propagation stage, we first use the SAM~3 detector to identify and segment same-category objects in each frame. We then extract object-level features for these candidate objects using DINOv3~\cite{simeoni2025dinov3} and match them against the target object from the first frame. The most similar candidates are retained as additional target prompts. During propagation, these selected masks, together with their corresponding frame indices, are encoded as initial memory, producing multiple high-quality anchors distributed across the video rather than only at the beginning. This design provides stronger guidance for the tracker in frames that are far from the first-frame prompt and helps prevent attention from drifting to distractor objects. With this simple re-prompting strategy, our method ranks {3rd} on the MOSEv2 test set and achieves {51.17} \JFo.

\section{Related Work}
\label{sec:relate}

\subsection{Semi-Supervised Video Object Segmentation}

Semi-supervised video object segmentation has recently been substantially advanced by SAM~2~\cite{ravi2024sam2}, which extends promptable segmentation from images to videos through a unified streaming-memory architecture. Given prompts such as points, boxes, or masks on one or multiple frames, SAM~2 propagates the target mask through the video by conditioning the current-frame representation on stored memories of previous prompts and predictions. Compared with earlier VOS systems designed for first-frame-mask propagation, SAM~2 unifies interactive video segmentation and semi-supervised VOS within a single framework, while also showing strong generalization across video domains. Its concise design and strong performance have made it the basis of many subsequent improvements.

Following SAM~2, a representative line of work focuses on improving memory construction and update. SAM2Long~\cite{ding2024sam2long} revisits the default memory management strategy in long videos and proposes more reliable memory selection for improved long-range propagation. SAMURAI~\cite{yang2024samurai} further introduces motion-aware mechanisms to reduce drift under fast motion and cluttered scenes. DAM4SAM~\cite{videnovic2025distractor} emphasizes distractor-aware memory design to better handle interference from similar objects, while SeC~\cite{zhang2025sec} augments fine-grained memory with higher-level concept cues from vision-language models. Although these methods differ in their specific designs, they largely share the same overall direction, namely strengthening SAM~2-style propagation by improving how memory is selected, updated, and interpreted.

More recently, SAM~3~\cite{sam3} extends the SAM~2 tracker with a parallel detector branch. In addition to promptable segmentation, SAM~3 supports concept-aware detection and segmentation based on text or visual exemplars, and combines detector outputs with tracker propagation to alleviate the gradual degradation that often appears in long videos. This design further broadens the framework toward more general video understanding and also provides a stronger basis for exploiting prompts beyond the first frame.

Despite these advances, existing methods still mainly improve \emph{how} memory is updated, selected, or enriched, while largely preserving the conventional inference paradigm of semi-supervised VOS: the first frame serves as the primary visual anchor, and the remaining frames are segmented by progressive propagation through memory-based association. Under the MOSEv2 setting, however, where targets frequently disappear and reappear and many same-category distractors coexist, relying on a single initial anchor remains fragile. This reveals a gap between stronger memory design and stronger target re-anchoring. Our method addresses this gap by introducing additional high-quality visual anchors in later frames through object-level retrieval and re-prompting, rather than relying solely on propagation from the first frame.

\subsection{Alternative Guidance Frames and Multi-Frame Anchoring}

\begin{figure*}[t]
    \centering
    \includegraphics[width=\textwidth]{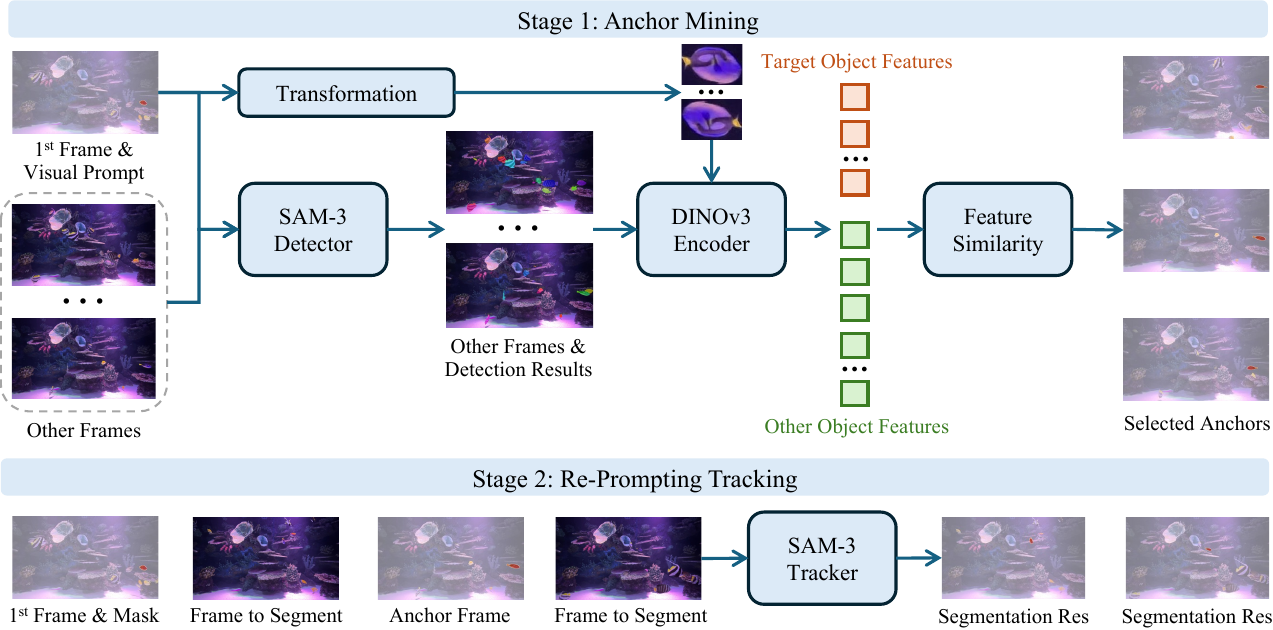}
    \caption{
    Overview of our two-stage framework. In the first stage, given the first-frame ground-truth mask as the visual prompt, we apply the SAM~3 detector to all remaining video frames to identify same-category object candidates, visualized by colored masks. Zoomed-in views are included for clearer visualization, as the target objects are often small. The detected regions are then encoded by DINOv3 and matched against a transformation-aware target feature pool constructed from the first-frame target, and a few high-confidence anchors are selected from later frames. In the second stage, the first-frame mask together with the selected anchor masks are used to re-prompt the SAM~3 tracker for mask propagation.
    }
    \label{fig:pipeline}
\end{figure*}

A related perspective in video object segmentation is that the first frame is not always the best or the only useful visual anchor. BubbleNets~\cite{griffin2019bubblenets} is an early representative work in this direction. Instead of fixing the first frame as the annotation frame, it studies how the choice of the guidance frame affects segmentation quality over the whole video and ranks candidate frames according to their expected utility for downstream propagation. Although its setting is different from modern SAM~based semi-supervised VOS, it suggests that the quality of the visual anchor can strongly affect later segmentation performance.

A more direct observation comes from XMem++~\cite{bekuzarov2023xmem++}, which shows that memory-based VOS can benefit from using multiple annotated frames rather than relying on a single initial reference. By retaining multiple guidance frames with diverse target appearances, it improves robustness under large pose changes, occlusion, and long videos. However, the additional frames in XMem++ are manually selected or iteratively provided by users, and the method is mainly developed for interactive settings.

This insight is also compatible with SAM~2~\cite{ravi2024sam2} and SAM~3~\cite{sam3}, whose prompting mechanisms naturally support masks, boxes, or points from multiple frames. Our method follows the same intuition that multiple anchors can improve segmentation, but differs in how these anchors are obtained. Instead of manually selecting additional frames, we automatically retrieve reliable target candidates from later frames using object-level similarity, and use the selected masks to re-prompt SAM~3. In this way, we turn multi-frame anchoring into an automatic procedure tailored to MOSEv2, where targets frequently disappear, reappear, and undergo severe transformations.

\section{Solution}

We address the MOSEv2 challenge by extending SAM~3 with an automatic re-prompting strategy based on object-level retrieval. The key idea is to move beyond the standard semi-supervised VOS setting where only the first frame serves as the visual anchor. Instead, we automatically identify reliable target candidates from later frames and use them as additional prompts, so that the tracker starts with multiple high-quality anchors distributed across the video. This design is particularly useful for MOSEv2, where the target frequently disappears and reappears, undergoes severe transformations, and is surrounded by many same-category distractors.

\subsection{Overview}

As illustrated in Fig.~\ref{fig:pipeline}, our framework consists of two consecutive components: automatic anchor mining and re-prompting-based tracking. Given the ground-truth mask of the target object in the first frame, conventional semi-supervised VOS methods directly use it as the only prompt and propagate the segmentation through the entire video. In contrast, our method first searches the video for reliable target instances that can serve as additional anchors, and then uses these anchors together with the first-frame mask to guide subsequent mask propagation. In this way, the tracker is supported not only by the initial frame, but also by several later frames that provide complementary target appearances.

\subsection{Automatic Anchor Mining and Re-Prompting}

Our method is built on SAM-3~\cite{sam3}, which contains both a tracker and a detector branch. The detector branch can identify and segment objects that belong to the same semantic category as a given visual prompt. We first use the provided target mask in the first frame as the initial visual prompt. Based on this prompt, the SAM-3 detector is applied to all remaining video frames to generate candidate objects of the same category as the target, as illustrated in Fig.~\ref{fig:pipeline}. For each candidate, the detector provides both a segmentation mask and the corresponding bounding box.

To distinguish the true target from same-category distractors, we further perform object-level matching using DINOv3 features~\cite{simeoni2025dinov3}. For the target object in the first frame, we crop the object region according to its mask and extract an object-level representation using DINOv3. We also construct transformed versions of the target object, including flipped and rotated variants, and extract DINOv3 features for these transformed views to form a small transformation-aware target feature pool. For each candidate object detected in later frames, we extract its object-level feature in the same way. We then compute the cosine similarity between each candidate and the target feature pool, and use the maximum similarity over the pool as the final matching score. This design improves robustness when the target undergoes large changes in pose, orientation, or appearance.

\begin{figure*}[t]
    \centering
    \includegraphics[width=\textwidth]{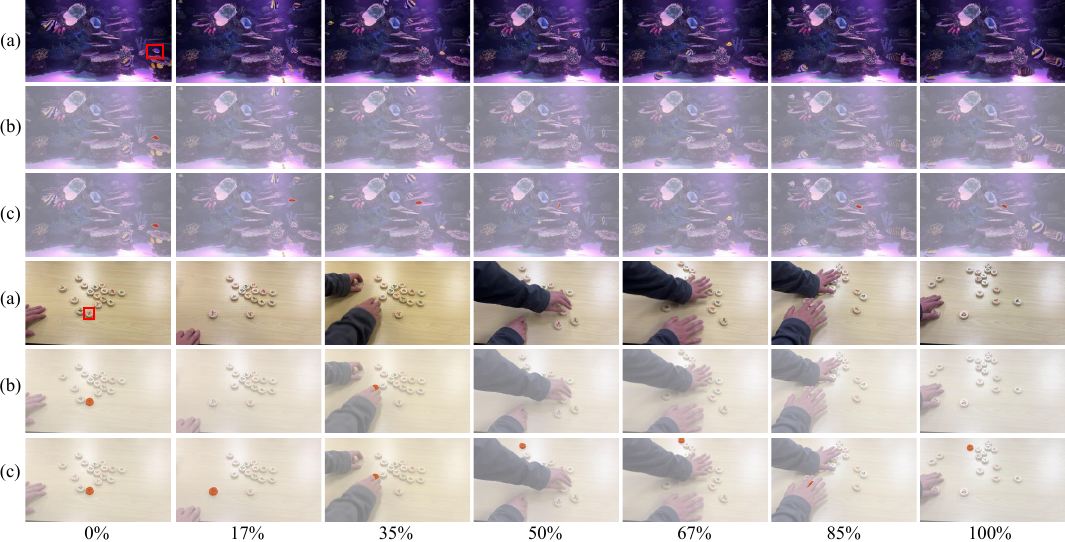}
    \caption{
    Qualitative comparison between first-frame-only propagation and our re-prompting strategy. The percentages indicate the temporal progress of each frame within the video. For each example, row (a) shows the original video frames, where the red box in the first frame marks the target object. Row (b) shows the result without re-prompting, where only the first-frame mask is used for propagation. Row (c) shows the result of our method.
    }
    \label{fig:qualitative}
\end{figure*}

After matching, we select a few high-confidence candidates as additional anchors. These selected masks are used as reliable pseudo-prompts that complement the original first-frame supervision. In our implementation, we retain top high-similarity candidates while suppressing temporally adjacent redundant selections, so that the final anchors are both reliable and temporally diverse.

Once these anchors are obtained, we inject them back into SAM~3 together with the first-frame ground-truth mask. Concretely, the first frame and the selected anchor frames are provided with masks, while the remaining video frames are fed into the SAM~3 tracker without masks, as shown in Fig.~\ref{fig:pipeline}. In this way, the tracker starts from multiple distributed anchors instead of a single one. Compared with standard first-frame-only propagation, this re-prompting strategy provides stronger target guidance in frames that are far from the initial prompt, and helps the tracker maintain the correct identity when severe transformations, long-term occlusion, disappearance and reappearance, or strong same-category distractors are present. As a result, the segmentation becomes more stable over long videos and less likely to drift to distractor objects.
\section{Experiments}

\subsection{Experimental Setup}

We train our method on the MOSEv2 training set for 10 epochs by fine-tuning all model parameters. The training is conducted on 8 NVIDIA A100 GPUs with a batch size of 1 and an input resolution of $1008$. We set the learning rate of the visual backbone to $3\times10^{-6}$ and use $5\times10^{-6}$ for the remaining parameters. Other training settings follow the default configuration of SAM~2.

During inference, we use an input resolution of $1288$. For automatic anchor selection, we rank candidate objects according to the object-level similarity score described in Sec.~3 and retain the top three high-similarity matches. To avoid selecting redundant anchors from temporally adjacent frames, we keep only one candidate among neighboring high-similarity responses. The selected masks are then used as additional prompts for re-prompting SAM~3.

\subsection{Main Results}

Table~\ref{tab:main_results} reports the final performance of our method on the MOSEv2 test set. Our method ranks {3rd} in the challenge leaderboard and achieves {51.17} on the official \JFo metric. This result suggests that automatically mining reliable anchors from later frames and injecting them back into the tracker is effective for handling long-range drift, distractor interference, and frequent disappearance-reappearance patterns in MOSEv2.

% ------------------------------------
\begin{table}[t]
\renewcommand{\arraystretch}{1.3}
\tabcolsep=0.1cm
\footnotesize
\centering
\begin{tabular}{p{0.2\columnwidth}|p{0.085\columnwidth}<{\centering}p{0.085\columnwidth}<{\centering}p{0.085\columnwidth}<{\centering}p{0.085\columnwidth}<{\centering}p{0.085\columnwidth}<{\centering}p{0.085\columnwidth}<{\centering}p{0.085\columnwidth}<{\centering}}
\toprule
Participant & \JFo & \J & \Fo & \JFnd & \JFnr & \F & \JF \\
\midrule
HITsz\_Dragon & 56.91 & 54.71 & 59.12 & 62.09 & 44.11 & 62.39 & 58.55 \\
tobedone & 55.16 & 53.02 & 57.29 & 58.51 & 44.47 & 61.02 & 57.02 \\
\rowcolor{blue!17!white}
HCVG & 51.17 & 49.55 & 52.80 & 66.23 & 32.60 & 55.13 & 52.34 \\
newsota & 48.19 & 46.16 & 50.22 & 55.36 & 37.21 & 52.97 & 49.56 \\
rozumrus & 44.83 & 43.18 & 46.48 & 55.81 & 31.29 & 48.96 & 46.07 \\
\rowcolor{gray!10}
\multicolumn{8}{c}{$\mathbf{\cdots}$} \\
\bottomrule
\end{tabular}
\caption{Quantitative comparisons on the MOSEv2 test set. Our results are highlighted in the blue row.}
\label{tab:main_results}
\end{table}
% ------------------------------------

\subsection{Qualitative Results}

We further present qualitative comparisons in Fig.~\ref{fig:qualitative}. As shown in the first example, after the target disappears and later reappears, our method can accurately recover and segment the correct target object. In contrast, the first-frame-only setting in row (b) either completely loses the target or remains continuously distracted by similar objects. This illustrates the benefit of introducing additional anchors beyond the first frame: even when the tracker reaches frames that are far from the initial prompt, the later anchors still provide effective target guidance and help prevent drift to distractors. A similar trend can also be observed in the second example, where re-prompting improves target recovery for a substantial portion of the video.

At the same time, our method is not always successful. In the second example, when the target reappears near the end of the video, it is still not recovered correctly. This suggests that, although the proposed training-free re-prompting strategy already improves robustness, the target representation can be further strengthened to better handle severe appearance and transformation changes.
\section{Conclusion}

In this report, we presented our solution to the MOSEv2 challenge based on automatic anchor mining and re-prompting. Starting from the SAM~3 framework, we introduced an object-level retrieval step to identify reliable target anchors from later frames, and used these anchors together with the first-frame mask to strengthen subsequent tracking. This design improves robustness to distractors, severe transformation, and target disappearance and reappearance, which are central challenges in MOSEv2. Our method ranks 3rd on the test set, showing that extending first-frame-only propagation with automatically discovered later anchors is an effective direction for challenging semi-supervised VOS. At the same time, the qualitative results also suggest room for improvement, especially in making the target representation more robust under extreme appearance changes.

{
    \small
    \bibliographystyle{ieeenat_fullname}
    \bibliography{main}

@String(PAMI = {IEEE Trans. Pattern Anal. Mach. Intell.})

@String(CVPR= {IEEE Conf. Comput. Vis. Pattern Recog.})

@String(ICCV= {Int. Conf. Comput. Vis.})

@String(TIP  = {IEEE Trans. Image Process.})

@String(ICLR = {Int. Conf. Learn. Represent.})

@String(PAMI  = {IEEE TPAMI})

@String(CVPR  = {CVPR})

@String(ICCV  = {ICCV})

@String(TIP   = {IEEE TIP})

@String(ICLR  = {ICLR})

@inproceedings{ding2023mose,
  title={MOSE: A new dataset for video object segmentation in complex scenes},
  author={Ding, Henghui and Liu, Chang and He, Shuting and Jiang, Xudong and Torr, Philip HS and Bai, Song},
  booktitle=ICCV,
  pages={20224--20234},
  year={2023}
}

@article{ding2025mosev2,
  title={MOSEv2: A more challenging dataset for video object segmentation in complex scenes},
  author={Ding, Henghui and Ying, Kaining and Liu, Chang and He, Shuting and Jiang, Xudong and Jiang, Yu-Gang and Torr, Philip HS and Bai, Song},
  journal={arXiv preprint arXiv:2508.05630},
  year={2025}
}

@article{gao2023deep,
  title={Deep learning for video object segmentation: a review},
  author={Gao, Mingqi and Zheng, Feng and Yu, James JQ and Shan, Caifeng and Ding, Guiguang and Han, Jungong},
  journal={Artificial Intelligence Review},
  volume={56},
  number={1},
  pages={457--531},
  year={2023},
  publisher={Springer}
}

@inproceedings{chen2025m,
  title={M\^{} 3-VOS: Multi-Phase, Multi-Transition, and Multi-Scenery Video Object Segmentation},
  author={Chen, Zixuan and Li, Jiaxin and Liang, Junxuan and Tan, Liming and Guo, Yejie and Lu, Cewu and Li, Yong-Lu},
  booktitle=CVPR,
  pages={29193--29202},
  year={2025}
}

@inproceedings{wang2025videodirector,
  title={Videodirector: Precise video editing via text-to-video models},
  author={Wang, Yukun and Wang, Longguang and Ma, Zhiyuan and Hu, Qibin and Xu, Kai and Guo, Yulan},
  booktitle=CVPR,
  pages={2589--2598},
  year={2025}
}

@article{xu2018youtube,
  title={Youtube-vos: A large-scale video object segmentation benchmark},
  author={Xu, Ning and Yang, Linjie and Fan, Yuchen and Yue, Dingcheng and Liang, Yuchen and Yang, Jianchao and Huang, Thomas},
  journal={arXiv preprint arXiv:1809.03327},
  year={2018}
}

@inproceedings{perazzi2016benchmark,
  title={A benchmark dataset and evaluation methodology for video object segmentation},
  author={Perazzi, Federico and Pont-Tuset, Jordi and McWilliams, Brian and Van Gool, Luc and Gross, Markus and Sorkine-Hornung, Alexander},
  booktitle=CVPR,
  pages={724--732},
  year={2016}
}

@inproceedings{oh2019video,
  title={Video object segmentation using space-time memory networks},
  author={Oh, Seoung Wug and Lee, Joon-Young and Xu, Ning and Kim, Seon Joo},
  booktitle=ICCV,
  pages={9226--9235},
  year={2019}
}

@inproceedings{ravi2024sam2,
  title={SAM 2: Segment Anything in Images and Videos},
  author={Ravi, Nikhila and Gabeur, Valentin and Hu, Yuan-Ting and Hu, Ronghang and Ryali, Chaitanya and Ma, Tengyu and Khedr, Haitham and R{\"a}dle, Roman and Rolland, Chloe and Gustafson, Laura and Mintun, Eric and Pan, Junting and Alwala, Kalyan Vasudev and Carion, Nicolas and Wu, Chao-Yuan and Girshick, Ross and Doll{\'a}r, Piotr and Feichtenhofer, Christoph},
  booktitle=ICLR,
  year={2025}
}

@article{yang2024samurai,
  title={SAMURAI: Motion-Aware Memory for Training-Free Visual Object Tracking with SAM 2},
  author={Yang, Cheng-Yeng and Huang, Hsiang-Wei and Jiang, Zhongyu and Chai, Wenhao and Hwang, Jenq-Neng},
  journal=TIP,
  year={2026},
  publisher={IEEE}
}

@inproceedings{ding2024sam2long,
  title={Sam2long: Enhancing sam 2 for long video segmentation with a training-free memory tree},
  author={Ding, Shuangrui and Qian, Rui and Dong, Xiaoyi and Zhang, Pan and Zang, Yuhang and Cao, Yuhang and Guo, Yuwei and Lin, Dahua and Wang, Jiaqi},
  booktitle=ICCV,
  year={2025}
}

@inproceedings{videnovic2025distractor,
  title={A distractor-aware memory for visual object tracking with sam2},
  author={Videnovic, Jovana and Lukezic, Alan and Kristan, Matej},
  booktitle=CVPR,
  pages={24255--24264},
  year={2025}
}

@inproceedings{zhang2025sec,
  title={SeC: Advancing Complex Video Object Segmentation via Progressive Concept Construction},
  author={Zhang, Zhixiong and Ding, Shuangrui and Dong, Xiaoyi and He, Songxin and Lin, Jianfan and Tang, Junsong and Zang, Yuhang and Cao, Yuhang and Lin, Dahua and Wang, Jiaqi},
  booktitle=ICLR,
  year={2025}
}

@article{ding2025mevis,
  title={MeViS: A multi-modal dataset for referring motion expression video segmentation},
  author={Ding, Henghui and Liu, Chang and He, Shuting and Ying, Kaining and Jiang, Xudong and Loy, Chen Change and Jiang, Yu-Gang},
  journal=PAMI,
  year={2025},
  publisher={IEEE}
}

@inproceedings{kirillov2023segment,
  title={Segment anything},
  author={Kirillov, Alexander and Mintun, Eric and Ravi, Nikhila and Mao, Hanzi and Rolland, Chloe and Gustafson, Laura and Xiao, Tete and Whitehead, Spencer and Berg, Alexander C and Lo, Wan-Yen and others},
  booktitle=ICCV,
  pages={4015--4026},
  year={2023}
}

@article{simeoni2025dinov3,
  title={Dinov3},
  author={Sim{\'e}oni, Oriane and Vo, Huy V and Seitzer, Maximilian and Baldassarre, Federico and Oquab, Maxime and Jose, Cijo and Khalidov, Vasil and Szafraniec, Marc and Yi, Seungeun and Ramamonjisoa, Micha{\"e}l and others},
  journal={arXiv preprint arXiv:2508.10104},
  year={2025}
}

@inproceedings{sam3,
      title={SAM 3: Segment Anything with Concepts},
      author={Nicolas Carion and Laura Gustafson and Yuan-Ting Hu and Shoubhik Debnath and Ronghang Hu and Didac Suris and Chaitanya Ryali and Kalyan Vasudev Alwala and Haitham Khedr and Andrew Huang and Jie Lei and Tengyu Ma and Baishan Guo and Arpit Kalla and Markus Marks and Joseph Greer and Meng Wang and Peize Sun and Roman Rädle and Triantafyllos Afouras and Effrosyni Mavroudi and Katherine Xu and Tsung-Han Wu and Yu Zhou and Liliane Momeni and Rishi Hazra and Shuangrui Ding and Sagar Vaze and Francois Porcher and Feng Li and Siyuan Li and Aishwarya Kamath and Ho Kei Cheng and Piotr Dollár and Nikhila Ravi and Kate Saenko and Pengchuan Zhang and Christoph Feichtenhofer},
      booktitle = ICLR,
      year = {2026}
}

@inproceedings{griffin2019bubblenets,
  title={Bubblenets: Learning to select the guidance frame in video object segmentation by deep sorting frames},
  author={Griffin, Brent A and Corso, Jason J},
  booktitle=CVPR,
  pages={8914--8923},
  year={2019}
}

@inproceedings{bekuzarov2023xmem++,
  title={Xmem++: Production-level video segmentation from few annotated frames},
  author={Bekuzarov, Maksym and Bermudez, Ariana and Lee, Joon-Young and Li, Hao},
  booktitle=ICCV,
  pages={635--644},
  year={2023}
}
}

\end{document}